\documentclass[11pt,a4paper]{article}
\usepackage[hyperref]{BLANCsensitivity}   
\usepackage{times}
\usepackage{latexsym}
\usepackage{graphicx} 
\usepackage{booktabs} 
\usepackage{placeins} 
\usepackage{microtype} 

\usepackage{microtype}

\aclfinalcopy 

\title{Sensitivity of BLANC to human-scored qualities of text summaries}

\author{Oleg Vasilyev, Vedant Dharnidharka, Nicholas Egan, Charlene Chambliss, John Bohannon \\
  Primer Technologies Inc. \\
  San Francisco, California \\
  \texttt{{oleg,vedant,negan,charlene,john}@primer.ai}\\}

\date{}

\begin{document}
\maketitle
\begin{abstract}
We explore the sensitivity of a document summary quality estimator, BLANC, to human assessment of qualities for the same summaries. In our human evaluations, we distinguish five summary qualities, defined by how fluent, understandable, informative, compact, and factually correct the summary is. We make the case for optimal BLANC parameters, at which the BLANC sensitivity to almost all of summary qualities is about as good as the sensitivity of a human annotator. 
\end{abstract}

\section{Introduction}

Human evaluation of summary quality \cite{Wojciech2019Neural} is arguably the most justified method for evaluating summarization algorithms, but it is neither consistent nor reproducible. Despite its drawbacks, human evaluation can be used to confirm the utility of any automated measure. The BLANC measure, introduced in \cite{Oleg2020Fill}, is an automated measure that is very semantic and yet easily reproducible. The aim of this paper is to understand how BLANC correlates with the human-assessed qualities of a summary, subject to the choice of BLANC parameters and to possible modifications of BLANC. For convenience of calculating BLANC we provide the python package blanc.

Human assessment can be favorably biased toward more extractive summaries \cite{Daniel2020FineTuning}, but this bias can be mitigated by splitting the overall `quality score' into multiple dimensions, such as fluency, informativeness, and factual correctness \cite{Stratos2019Sum, Wojciech2019Neural, Wojciech2017Evaluating, Lisa2018Robust, Zhu2020Boosting}. We considered well defined dimensions in our human evaluations, and in this paper we consider how sensitive BLANC is to each of them. We also propose what we see as the `optimal' usage of BLANC.

\section{BLANC: an objective, reproducible measure}

BLANC is defined as a measure of how well a summary helps an independently trained model while the model performs a language understanding task on a document \cite{Oleg2020Fill}. Both the model and the language understanding task are expected to be unrelated to the summarization task. The model is assumed to be trained on generic texts.

More specifically, the well-known trained BERT-base language model \cite{Jacob2018BERT} and the Cloze task \cite{Wilson1953Cloze} were used in \cite{Oleg2020Fill}. The model reconstructs masked tokens, where the tokens are words or parts of words as defined by the BERT model dictionary. BLANC is defined as follows:

\[BLANC = \frac{N_{help} - N_{base}}{N_{total}}\]

Here $N_{help}$ is the number of successfully unmasked tokens when the summary is used by the model; $N_{base}$ is the number of successfully unmasked tokens when the summary is not used by the model; $N_{total}$ is the total number of masked tokens. The tokens are masked evenly, every $M$th token being masked, with optional restrictions as specified in the pseudo-code in Figure \ref{fig:Algo_help}. Figure \ref{fig:Algo_help} corresponds to BLANC version that was called `BLANC-help' in \cite{Oleg2020Fill}.

 \begin{figure}[htb]
     \centering
     \begin{tabular}{l}
     \toprule
     Given: $summary$; $text$; $model$;\\ 
     \hspace{.4cm}$M=6$, $L_w=4$, $L_s=0$, $L=1000$\\  \\
     
     {\bf{def}} $mask\_token(tok, i)$:\\
        \hspace{.4cm}if (tok is word and $len(tok)<L_w$ or\\
            \hspace{.4cm}\hspace{.4cm}tok starts word and $len(tok)<L_s$ or\\
            \hspace{.4cm}\hspace{.4cm}tok continues word and $len(tok)<L$ or\\
            \hspace{.4cm}\hspace{.4cm}$i\%M \neq 0$):\\
            \hspace{.4cm}\hspace{.4cm}return\\
        \hspace{.4cm}tok = mask\\ \\
     
     Initialise $filler = "." * length(summary)$\\
     $N_{base}=0$, $N_{help}=0$, $N_{total}=0$\\
     \bf{for} $sentence$ in $text$:\\
     \hspace{.4cm}\bf{for} $i_{0}$ in range from $1$ to $M$:\\
     \hspace{.4cm}\hspace{.4cm}{\bf{for each}} $i$th $token$ in $sentence$:\\
     \hspace{.4cm}\hspace{.4cm}\hspace{.4cm}$mask\_token(token, i-i_0)$\\
     \hspace{.4cm}\hspace{.4cm}$input_{base} = filler + sentence$\\
     \hspace{.4cm}\hspace{.4cm}$input_{help} = summary + sentence$\\
     \hspace{.4cm}\hspace{.4cm}$prediction_{base} = model(input_{base})$\\
     \hspace{.4cm}\hspace{.4cm}$prediction_{help} = model(input_{help})$\\
     \hspace{.4cm}\hspace{.4cm}{\bf{for each}} position $i$ in masked tokens:\\
     \hspace{.4cm}\hspace{.4cm}\hspace{.4cm}if $prediction_{base}[i] == sentence[i]$:\\
        \hspace{.4cm}\hspace{.4cm}\hspace{.4cm}\hspace{.4cm}$N_{base} += 1$\\
     \hspace{.4cm}\hspace{.4cm}\hspace{.4cm}if $prediction_{help}[i] == sentence[i]$:\\
        \hspace{.4cm}\hspace{.4cm}\hspace{.4cm}\hspace{.4cm}$N_{help} += 1$\\
     \hspace{.4cm}\hspace{.4cm}\hspace{.4cm}$N_{total} += 1$\\
     $B = (N_{help} - N_{base}) / N_{total}$\\
     \bottomrule
     \end{tabular}
     \caption{BLANC score $B$ for quality of summary.}
     \label{fig:Algo_help}
 \end{figure}
 
Note that the algorithm is simple, and the trained BERT model is available as `BertForMaskedLM' in transformers \cite{huggingface2019website}. For convenience, we provide a python package github.com/PrimerAI/blanc for calculating BLANC. 

The BLANC score is affected by several parameters. The settings chosen in \cite{Oleg2020Fill} are grounded in commonsense intuition. The threshold $L_{w}$ for the length of normal tokens (a token representing a whole word) is set to be 4 characters, meaning that the shorter tokens are not masked. Short tokens are more likely to be stopwords, which would be easily guessed by the model irrespective of any help from the summary. If a word is split, the start token is always masked, because words that were not found in the model dictionary are likely to be more informative. For split words, tokens other than the start token are never masked. The $M=6$ demands masking every 6th token (~16.7\% of tokens) because this choice naturally corresponds to the BERT model's training masking frequency of 15\% of tokens \cite{Jacob2018BERT}.

With these parameters, the BLANC values for summaries generated by popular summary generation methods usually range from 0.05 to 0.20 (and sometimes up to 0.40), meaning that a summary usually provides an additional 5\%-20\% to the model's ability to guess masked tokens from the original document.

How can we can validate possible alternative choices in parameters for BLANC? Empirically, we want to have higher correlations between BLANC scores and human assessments, and a higher average value of BLANC over a given set of generated summaries (implying that BLANC is more sensitive to the summaries). 

\section{Best frequency of masking}
\label{sec:best_freq}
We have found that alternative settings result in either negligible changes to the final BLANC scores, or in lowered correlations between BLANC and human scores. The only exception is that decreasing the masking step $M$ does seem to increase the correlation with human scores. 

To illustrate this finding, we provide the results of BLANC's correlation with the human scores obtained in our human evaluation.

The summaries used in the human evaluation were generated by algorithms generating summaries in three distinctly different styles: Microsoft's abstractive UniML model \cite{Li2019Unified}, a semi-abstractive model \cite{oleg2019headline}, and the purely extractive LexRank model \cite{Gunes2004LexRank}. We assembled 300 summary-text pairs for human scoring, where the 100 texts were randomly sampled news articles. We hired 10 annotators through Odetta.ai and trained them to assess five qualities of each summary, each quality on a 5-point scale: 0 = VERY BAD, 1 = BAD, 2 = OK, 3 = GOOD or 4 = VERY GOOD. The annotators worked independently from each other and had access to only one summary-text pair at a time, to avoid biasing their judgments. The task was performed through the online text annotation tool LightTag (lighttag.io).

Comprehensive instructions were given to the annotators, including examples for each quality. Excerpts of the description for each quality are given below:

\begin{enumerate}
\item $Fluent$: For fluency, the only consideration is whether the summary is easy to read.
\item $Understandable$: A summary is “understandable” if a reader can confidently and unambiguously restate the facts that were communicated in the summary.
\item $Informative$: A summary is “informative” if it contains all of the “most important” information from the article.
\item $Compact$: A summary is “compact” if the summary does not contain irrelevant or unimportant information. 
\item $Overall$: The “Overall” quality should reflect how good a summary is in general. The judgement should be based on all your ratings of the specific qualities, and can also take into account any obviously noticeable factual errors. (We asked you to ignore factual errors while rating the previous qualities, but you may account them in your rating for Overall.)
\end{enumerate}

Factual correctness was evaluated separately and will be covered in the next section. The first two qualities - Fluency and Understadability - do not require reading the text. It is even better if while evaluating these qualities the annotators read only the summary, thus avoiding an influence of the other qualities (Informativeness, Compactness, and Factual correctness).

In defining the qualities, our goal was to make scoring as unambiguous as possible, and the definitions above result from iterating and refining the definitions with feedback from annotators. The qualities we ultimately decided on were in part inspired by previous works on summary evaluation. For example, in \cite{Stratos2019Sum} the `grammaticality', `structure \& coherence' and `referential clarity' correspond to what we call `fluency'; the `focus' corresponds to our `informativeness'; the `non redundancy' partially to our `compactness'; and two of our qualities (understandability and factual correctness) were not addressed. In \cite{Wojciech2019Neural} `relevance' corresponds to a combination of what we call `informativeness' and `compactness'; `consistency' corresponds to factual correctness; `fluency' and `coherence' roughly correspond to our fluency, but we do not separate intra- and inter-sentence fluency; lastly, our quality of understandability is not covered. Four dimensions used in \cite{Lisa2018Robust} are called `non-redundancy', `fluency', `faithfulness to input' and `informativeness', -- they can also be partially mapped to our qualities.  

\begin{figure*}[!htbp]
    \raggedleft
    \includegraphics[width=16.0cm]{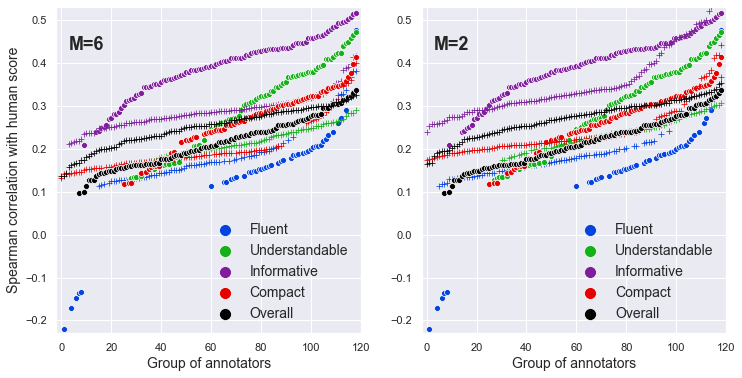}
    \caption{Spearman correlations of BLANC and human scores with a group of 7 annotators. Step $M=6$ left, and $M=2$ right. Note that the x-axis is not a measure; it represents all 120 ways $10 \choose 3$ to split 10 annotators into groups of 3 and 7. Circle-markers represent the correlation between annotator groups: avg. score of the group of 3 correlated with the avg. score of the group of 7. Plus-markers represent the correlation between the BLANC score and the group of 7 annotators. Each type of correlation is sorted independently, left-to-right. Correlation coefficients that were not significant $(p < 0.05)$ are not shown.}
    \label{fig:blanc_qualities_gap6and2}
\end{figure*}

Each text-summary pair was scored by 10 annotators. The BLANC score was also calculated on each pair. The Figure \ref{fig:blanc_qualities_gap6and2} illustrates a simple comparison human assessments with BLANC with mask steps $M=6$ and $M=2$. 

To compare BLANC with human assessments, we split our 10 annotators into one group of 3 and a second group of the remaining 7. There are 120 ways to make this split $10 \choose 3$, hence there are 120 groups of annotators on the X-axis. The circle markers show the human-human correlation: the correlation between the average score of the small group and the average score of the large group. The plus markers show BLANC-human correlation: the correlation between BLANC and the large group of annotators. Thus, the comparison being made is between BLANC and a 'team' of 3 annotators. For simplicity of the presentation, each type of correlation was sorted independently. Correlation coefficients that were not significant $(p < 0.05)$ are not shown.

The figures show that BLANC has stronger agreement with the large group on the `fluent' and `overall' qualities, but the small human group is stronger on `understandable', `compact', and especially on `informative'. We also see that BLANC becomes more strong when $M=2$, even for the `informative' quality.

\begin{figure}[!htb]
    \centering
    \includegraphics[width=7.7cm]{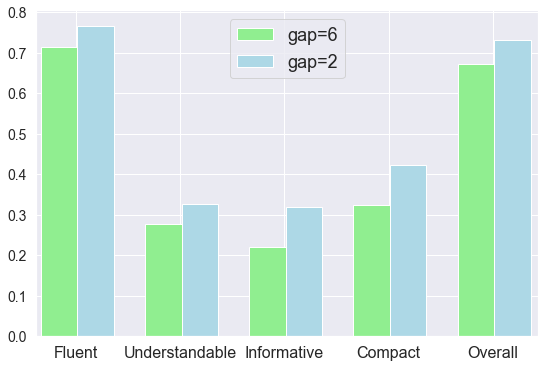}
    \caption{Height of the bar shows a fraction of cases in which the Spearman correlation of the `other 7' annotators was higher with BLANC than with 3 annotators.}
    \label{fig:blanc_human_gap26_Spearman}
\end{figure}

The same trend can be seen in Figure \ref{fig:blanc_human_gap26_Spearman}, in which we count the fraction of instances in which BLANC outperformed the 3-annotators group in its correlation with the 7-annotators group. This comparison also shows that BLANC outperforms the group of 3 humans more often when $M=2$.

Since the correlation of BLANC with human assessments increased monotonically when decreasing the step $M$ from 6 to 2, we attempted to go even further to lower `step' (more frequent masking), by masking two or more neighboring tokens rather than one. This did not lead to better correlations. Thus, we can say that $M=2$ (masking exactly half of tokens) seems to be the ideal choice for producing high-performing BLANC scores. BLANC's average score across all summaries is also highest at $M=2$, implying that on average BLANC extracts maximal help from a summary at $M=2$. 

\section{Factual errors}
The remaining quality of a summary is factual correctness. It is arguably much more objective than other qualities; reasonable people may disagree on what makes for a good summary, but factual errors are binary: a statement implies an erroneous conclusion, or it doesn't. 

We described ``factual correctness" as follows to the annotators: ``Factual correctness'' is only about whether or not the information in the summary is correct. We are not concerned with grammar, spelling, repetition, or how readable the summary is. 

We used the following classifications for different types of errors (excerpts are taken from the instructions for our annotators):
\begin{enumerate}
\item {\it{Incorrect named entity}}: The summary includes the incorrect person, organization, location or some other named entity.
\item {\it{Incorrect data}}: The summary describes an event with an incorrect date, time, number, or some other data.
\item {\it{Cascading error}}: It can happen that an incorrect named entity error (INE) affects one or more sentences down in the summary. If an incorrect sentence can be made correct by modifying one of the previous sentences, then it is a cascading error. If not, it is not a cascading error.
\item {\it{Hallucination}}: The summary includes a person, event, fact, or quote that does not show up anywhere in the article. 
\item {\it{Negation}}: The summary says the opposite of what the article says.
\item {\it{Other}}: Assign this tag to anything else that is a factual error but that is not assigned to one of the above categories. Remember that a factual error must be a definitively wrong statement, not a statement that is confusing or unclear. 
\end{enumerate}
Additional ``assign by origin" rule was given for distinction between incorrect named entity and incorrect data: the error must be defined by “what was supposed to be there,” rather than “what the summary wrote there instead.”

The factual correctness task is more difficult and time consuming than assessment of the qualities considered in the previous section. Using the same dataset of 300 generated summaries, we asked three annotators, after an initial training round, to complete the task. 

We observed that each annotated error was annotated by only slightly more than one annotator on average, and each annotated (with one or more errors) summary was annotated by almost two annotators. After the reconciliation (ensuring all annotators agreed on each annotation), the end result was a total of 65 factual errors in 44 erroneous summaries (out of 300 summaries). The most frequent error in all kinds of summaries (extractive, semi-abstractive and abstractive) is the incorrect named entity. The abstractive summaries have almost as much hallucination errors, and in total had the most errors. The extractive summaries have the least errors.

The evaluation showed that BLANC is not very sensitive to factual errors. The Spearman correlation between BLANC with $M = 2$ and number of error-annotators per summary was -0.14 with a p-value of 0.02. The Pearson correlation was -0.13 with a p-value of 0.03. The correlation sign is correct: errors make summary worse, hence the negative correlation. But the correlation is fairly weak.

We verified our conclusion by using synthetic data: we simulated entity-swap errors on human summaries of the CNN / Daily Mail dataset \cite{Karl2015Teaching}. We selected randomly 1500 Daily Mail summaries, and 1500 CNN summaries. The errors were introduced through entity swapping \cite{Wojciech2017Evaluating}, using github.com/salesforce/factCC. 

If BLANC responded perfectly to simulated errors,  we would expect its value to always decrease when entities are swapped within a summary. In reality, for BLANC with mask step $M = 2$, this only happened in 56\% of cases, and the opposite happened in 32\% of cases. This trend did not change when the mask step size was increased to 6. CNN summaries produced similarly poor results: 50\% of scores decreased in response to the error, and 37\% increased. A sum of squares of BLANC with mask steps 2 and 6 on Daily Mail summaries gives 60\% against 33\%. As the most challenging quality for BLANC to assess, it is clear that factual correctness continues to present challenges for any automated summary evaluation algorithm.

\section{Alternatives to accuracy in the Cloze task}

In this paper, we used the original definition of BLANC, which measures the amount by which a summary increases model accuracy on the Cloze task. This  choice was motivated by analogy with how a summary might assist a human: helping a reader understand a text in a way that would help her reconstruct missing portions of the original text. But since the model also produces a distribution over possible tokens for every masked token, one could form alternative BLANC metrics based on the predicted logit or probability of the correct token.   

We can define a generalized BLANC score as
$$ BLANC = \frac{1}{N_{total}} \sum_{i=1}^{N_{total}} (x_s^{(i)} - x_f^{(i)}) $$
where $N_{total}$ is the number of masked tokens, $x_s^{(i)}$ is the logit, probability, or log probability of the correct token being unmasked for masked token $i$ when using the summary, and $x_f^{(i)}$ corresponds to using the filler instead of the summary.

We compare the `performance' of using the logit, probability, and log probability for BLANC against the original BLANC formulation in Tables \ref{tab:blanc_alternative_news} and \ref{tab:blanc_alternative_dmcnn}. Performance is measured by correlation with overall human scores (not split into qualities). The table \ref{tab:blanc_alternative_news} shows results for the same dataset that we used in the section \ref{sec:best_freq}: 300 text-summary pairs, with 3 summaries generated for each of 100 random news texts. The table \ref{tab:blanc_alternative_dmcnn} does the same for 300 summaries generated from 100 random texts taken from the DM/CNN dataset \cite{Karl2015Teaching}. In these experiments, we used a mask step of 2. The results show that using the logits or probabilities is a promising alternative that would benefit from further exploration.

\begin{table}[!htb]
\centering
\caption{A comparison of alternative BLANC metrics to the performance of standard BLANC, as measured by correlation with human scores for a sample of 300 summaries generated for 100 random news documents.}
\label{tab:blanc_alternative_news}
\begin{tabular}{|l|r|r|} 
\hline
\textbf{Metric} & \multicolumn{1}{l|}{\textbf{Pearson}} & \multicolumn{1}{l|}{\textbf{Spearman}}  \\ 
\hline
Original        & 0.372                                 & 0.343                                   \\ 
\hline
Logits          & 0.367                                 & 0.349                                   \\ 
\hline
Probabilities   & 0.370                                 & 0.344                                   \\ 
\hline
Log Probs       & 0.317                                 & 0.287                                   \\
\hline
\end{tabular}
\end{table}
\begin{table}
\centering
\caption{A comparison of alternative BLANC metrics to the performance of standard BLANC, as measured by correlation with human scores for a sample of 300 summaries generated for 100 random documents from the CNN/Daily Mail dataset.}
\label{tab:blanc_alternative_dmcnn}
\begin{tabular}{|l|r|r|} 
\hline
\textbf{Metric} & \multicolumn{1}{l|}{\textbf{Pearson}} & \multicolumn{1}{l|}{\textbf{Spearman}}  \\ 
\hline
Original        & 0.183                                 & 0.174                                   \\ 
\hline
Logits          & 0.222                                 & 0.189                                   \\ 
\hline
Probabilities   & 0.215                                 & 0.199                                   \\ 
\hline
Log Prob        & 0.178                                 & 0.168                                   \\
\hline
\end{tabular}
\end{table}
\section{Conclusion}

In this paper we presented our results on the sensitivity of BLANC, an automated measure for the assessment of summary quality, to several different dimensions of summary quality as assessed by humans. The qualities considered included fluency, clarity, informativeness, compactness and factual correctness, as well as an ``overall" judgment of the summary's quality. 

We found that BLANC is reasonably sensitive to all the qualities, but least of all to factual correctness. We observed that BLANC is especially sensitive to overall quality and to fluency, producing superior assessments when compared to a group of 3 human annotators. We have found optimal parameters for BLANC. We also have shown that several reasonable variations of BLANC do not lead to substantially better results, and presented avenues for future directions for improving the performance of BLANC.

\bibliography{BLANCsensitivity}

\begin{thebibliography}{14}
\expandafter\ifx\csname natexlab\endcsname\relax\def\natexlab#1{#1}\fi

\bibitem[{Devlin et~al.(2018)Devlin, Chang, Lee, and Toutanova}]{Jacob2018BERT}
Jacob Devlin, Ming-Wei Chang, Kenton Lee, and Kristina Toutanova. 2018.
\newblock \href {http://arxiv.org/abs/1810.04805} {Bert: Pre-training of deep
  bidirectional transformers for language understanding.}
\newblock \emph{arXiv}, arXiv:1810.04805.

\bibitem[{Dong et~al.(2019)Dong, Yang, Wang, Wei, Liu, Wang, Gao, Zhou, and
  Hon}]{Li2019Unified}
Li~Dong, Nan Yang, Wenhui Wang, Furu Wei, Xiaodong Liu, Yu~Wang, Jianfeng Gao,
  Ming Zhou, and Hsiao-Wuen Hon. 2019.
\newblock \href {http://arxiv.org/abs/1905.03197} {Unified language model
  pre-training for natural language understanding and generation.}
\newblock \emph{arXiv}, arXiv:1905.03197.

\bibitem[{Erkan and Erkan(2004)}]{Gunes2004LexRank}
Güneş Erkan and Günes Erkan. 2004.
\newblock Lexrank: Graph-based centrality as salience in text summarization.
\newblock \emph{Journal of Artificial Intelligence Research}, 22(1):457--479.

\bibitem[{Fan et~al.(2018)Fan, Yu, and Wang}]{Lisa2018Robust}
Lisa Fan, Dong Yu, and Lu~Wang. 2018.
\newblock \href {http://arxiv.org/abs/1810.06065} {Robust neural abstractive
  summarization systems and evaluation against adversarial information.}
\newblock \emph{arXiv}, arXiv:1810.06065.

\bibitem[{Hermann et~al.(2015)Hermann, Kočiský, Grefenstette, Espeholt, Kay,
  Suleyman, and Blunsom}]{Karl2015Teaching}
Karl~Moritz Hermann, Tomáš Kočiský, Edward Grefenstette, Lasse Espeholt,
  Will Kay, Mustafa Suleyman, and Phil Blunsom. 2015.
\newblock Teaching machines to read and comprehend.
\newblock In \emph{Advances in Neural Information Processing Systems 28}, pages
  1693--1701. Curran Associates, Inc.

\bibitem[{Kryściński et~al.(2019{\natexlab{a}})Kryściński, Keskar, McCann,
  Xiong, and Socher}]{Wojciech2019Neural}
Wojciech Kryściński, Nitish~Shirish Keskar, Bryan McCann, Caiming Xiong, and
  Richard Socher. 2019{\natexlab{a}}.
\newblock Neural text summarization: A critical evaluation.
\newblock In \emph{Proceedings of the 2019 Conference on Empirical Methods in
  Natural Language Processing and the 9th International Joint Conference on
  Natural Language Processing (EMNLP-IJCNLP)}, pages 540--551. Association for
  Computational Linguistics.

\bibitem[{Kryściński et~al.(2019{\natexlab{b}})Kryściński, McCann, Xiong,
  and Socher}]{Wojciech2017Evaluating}
Wojciech Kryściński, Bryan McCann, Caiming Xiong, and Richard Socher.
  2019{\natexlab{b}}.
\newblock \href {http://arxiv.org/abs/1910.12840} {Evaluating the factual
  consistency of abstractive text summarization.}
\newblock \emph{arXiv}, arXiv:1910.12840.

\bibitem[{Taylor(1953)}]{Wilson1953Cloze}
Wilson~L Taylor. 1953.
\newblock Cloze procedure: A new tool for measuring readability.
\newblock \emph{Journalism Bulletin}, 30(4):415--433.

\bibitem[{Vasilyev et~al.(2020)Vasilyev, Dharnidharka, and
  Bohannon}]{Oleg2020Fill}
Oleg Vasilyev, Vedant Dharnidharka, and John Bohannon. 2020.
\newblock \href {http://arxiv.org/abs/2002.09836} {Fill in the blanc:
  Human-free quality estimation of document summaries.}
\newblock \emph{arXiv}, arXiv:2002.09836.
\newblock To appear in Proc. 1st Workshop on Evaluation and Comparison for NLP
  systems, 2020.

\bibitem[{Vasilyev et~al.(2019)Vasilyev, Grek, and Bohannon}]{oleg2019headline}
Oleg Vasilyev, Tom Grek, and John Bohannon. 2019.
\newblock \href {http://arxiv.org/abs/1904.08455v3} {Headline generation:
  Learning from decomposable document titles.}
\newblock \emph{arXiv}, arXiv:1904.08455v3.

\bibitem[{Wolf et~al.(2019)Wolf, Debut, Sanh, Chaumond, Delangue, Moi, Cistac,
  Rault, Louf, Funtowicz, and Brew}]{huggingface2019website}
Thomas Wolf, Lysandre Debut, Victor Sanh, Julien Chaumond, Clement Delangue,
  Anthony Moi, Pierric Cistac, Tim Rault, Rémi Louf, Morgan Funtowicz, and
  Jamie Brew. 2019.
\newblock \href {http://arxiv.org/abs/1910.03771} {Huggingface's transformers:
  State-of-the-art natural language processing.}
\newblock \emph{arXiv}, arXiv:1910.03771.

\bibitem[{Xenouleas et~al.(2019)Xenouleas, Malakasiotis, Apidianaki, and
  Androutsopoulos}]{Stratos2019Sum}
Stratos Xenouleas, Prodromos Malakasiotis, Marianna Apidianaki, and Ion
  Androutsopoulos. 2019.
\newblock Sumqe: a bert-based summary quality estimation model.
\newblock In \emph{Proceedings of the 2019 Conference on Empirical Methods in
  Natural Language Processing and the 9th International Joint Conference on
  Natural Language Processing (EMNLP-IJCNLP)}, pages 6005--6011, Hong Kong,
  China. Association for Computational Linguistics.

\bibitem[{Zhu et~al.(2020)Zhu, Hinthorn, Xu, Zeng, Zeng, Huang, and
  Jiang}]{Zhu2020Boosting}
Chenguang Zhu, William Hinthorn, Ruochen Xu, Qingkai Zeng, Michael Zeng,
  Xuedong Huang, and Meng Jiang. 2020.
\newblock \href {http://arxiv.org/abs/2003.08612v4} {Boosting factual
  correctness of abstractive summarization.}
\newblock \emph{arXiv}, arXiv:2003.08612v4.

\bibitem[{Ziegler et~al.(2020)Ziegler, Stiennon, Wu, Brown, Radford, Amodei,
  Christiano, and Irving}]{Daniel2020FineTuning}
Daniel~M. Ziegler, Nisan Stiennon, Jeffrey Wu, Tom~B. Brown, Alec Radford,
  Dario Amodei, Paul Christiano, and Geoffrey Irving. 2020.
\newblock \href {http://arxiv.org/abs/1909.08593v2} {Fine-tuning language
  models from human preferences.}
\newblock \emph{arXiv}, arXiv:1909.08593v2.

\end{thebibliography}
\bibliographystyle{acl_natbib}

\appendix

\end{document}